\documentclass{article}

\usepackage{natbib}
\usepackage{PRIMEarxiv}
\usepackage{tikz}
\usetikzlibrary{arrows.meta,positioning,fit,calc}

\usepackage[utf8]{inputenc} %
\usepackage[T1]{fontenc}    %
\usepackage{hyperref}       %
\usepackage{url}            %
\usepackage{booktabs}       %
\usepackage{amsfonts}       %
\usepackage{nicefrac}       %
\usepackage{microtype}      %
\usepackage{lipsum}
\usepackage{fancyhdr}       %
\usepackage{graphicx}       %
\graphicspath{{media/}}     %

\usepackage{amsmath}
\usepackage{tabularx}
\usepackage{todonotes}
\usepackage{multirow}
\usepackage{algorithm}
\usepackage{algpseudocode}

\title{Thinking Deeper, Not Longer: Memory-Efficient Latent Chain-of-Thought with Depth-Recurrent Transformers for Compositional Generalization}

\author{
  Hung-Hsuan Chen \\
  Computer Science and Information Engineering \\
  National Central University \\
  Taoyuan, Taiwan \\
  \texttt{hhchen1105@acm.org}
}

\begin{document}
\maketitle

\begin{abstract}

Standard Transformers have a fixed computational depth, limiting their ability to generalize to tasks that require variable-depth reasoning. The usual remedy, Chain-of-Thought (CoT), spends tokens to reason, inflating the key--value cache, so memory becomes the limiting cost when reasoning is served over large query batches. We study a depth-recurrent Transformer that decouples computational depth from parameter count by iterating a shared-weight block, so that adding reasoning steps leaves peak inference memory unchanged and grows latency only linearly, with no token generation. Three ingredients support the recurrence over 20+ thinking steps: a silent thinking objective that supervises only the final output, LayerScale initialization, and an identity-biased gate that opens a gradient highway across steps. We characterize it on three compositional domains with decreasing structural bias: graph reachability (adjacency masking), nested boolean logic (relative positioning), and unstructured relational text (no structural positional cue). We find a \emph{computational frontier}: accuracy climbs once the thinking-step count meets the task's complexity, reaching near-perfect performance on the two structured tasks and a lower plateau on unstructured text. How it climbs depends on the structural bias---abruptly from chance on the graph task, gradually on the other two. Depth recurrence extrapolates beyond the training range: it succeeds on the graph task where fixed-depth models barely extrapolate, and on the two sequence tasks comes within two points of fixed-depth Transformers that use $4$--$6.4\times$ more parameters. On the graph task, whose adjacency mask makes propagation depth verifiable, intermediate per-step supervision---a standard recipe for deep iterative models---consistently \emph{harms} this extrapolation at widths $d \geq 256$. We release the code for reproducibility.\footnote{\url{https://github.com/hhchen1105/Depth-Recurrent-Transformer}}

\end{abstract}


\section{Introduction}\label{sec:intro}

Large Language Models (LLMs) have achieved remarkable performance across tasks. However, their reasoning capabilities remain constrained by their fixed-depth architecture. When faced with problems that require multi-step logical deduction, such as planning or mathematical proof, current models rely on \emph{Chain-of-Thought} (CoT)~\citep{wei2022chain}, which externalizes reasoning as a sequence of generated tokens. We call this paradigm \emph{horizontal} CoT: the model ``thinks'' by extending the output sequence horizontally.
 
Although horizontal CoT is effective, it suffers from the following limitations. First, each reasoning step costs tokens, consuming the finite context window. Second, generating intermediate reasoning as discrete tokens introduces a distinct failure mode: each sampled token is a discrete symbol drawn from a finite vocabulary, collapsing the model's continuous internal representation into a single categorical choice that all subsequent tokens must condition on. A continuous latent vector carries no such limitation: each entry remains real-valued, so later computation may reweight or attenuate an early error. Moreover, when horizontal CoT is trained with per-token supervision, the model learns to imitate each intermediate step in the training data rather than to reach the correct final answer.
 
In this work, we study an orthogonal paradigm that we call \emph{vertical} Chain-of-Thought: instead of generating tokens horizontally, the model ``thinks deeper'' by recurrently applying a shared-weight Transformer block. Vertical CoT is thus a form of \emph{latent} Chain-of-Thought: the intermediate reasoning stays in continuous hidden states rather than discrete tokens, but unlike prior latent approaches that still feed continuous thoughts back as additional sequence positions~\citep{hao2024training}, it unrolls along depth and leaves the sequence length unchanged. This approach decouples computational depth from both parameter count and context length. The model can invest more computation in harder instances by increasing the recurrence steps during inference. Because no tokens are emitted, the key--value cache does not grow with reasoning depth. On a depth-$24$ backbone, the cache for $64$ thinking tokens alone measures $2.1$--$3.3$ times our entire recurrent core, and it keeps growing with each further step while our core's footprint stays fixed (Section~\ref{sec:complexity}).
 
A practical concern in making depth recurrence work is stability. Unrolling a single shared-weight block for many steps is classically prone to exploding or vanishing gradients~\citep{pascanu2013difficulty} and representation collapse~\citep{dong2021attention}. However, a plain weight-tied block reaches competitive accuracy in our tasks (Section~\ref{sec:baselines}); what it lacks is stability: in one of our tasks (graph reachability, without the gate), out-of-distribution (OOD) accuracy ranges from $62$ to $76$ across random seeds. We obtain more reliable optimization and better-behaved gradients from three complementary mechanisms (Section~\ref{sec:method}):
 
\begin{enumerate}
    \item Silent Thinking: We apply supervision \emph{only} at the final recurrence step, with no intermediate auxiliary losses. This forces the model to develop genuine multi-step reasoning paths rather than learning heuristic shortcuts.
 
    \item LayerScale Initialization: Per-channel scaling initialized at $10^{-4}$ after attention and feed-forward layers, preventing initial random weights from corrupting preserved hidden states during early training.

    \item Identity-Biased Recurrence: To improve gradient propagation through the up to $16$ recurrence steps unrolled in training, we use a gated recurrence with the gate bias initialized to $-2.0$, strongly biased toward preserving the previous hidden state. This creates a gradient highway across recurrence steps (analyzed in Section~\ref{sec:gate-gradient}).
\end{enumerate}

To evaluate the Depth-Recurrent Transformer's capabilities, we design three OOD reasoning tasks. We begin with \emph{Graph Reachability}, which provides a physically verifiable testbed using topological masking. We then increase the structural complexity with \emph{Nested Boolean Logic}, demonstrating the model's ability to maintain hierarchical states using relative positioning. Finally, we remove all task-aligned structural inductive biases in a \emph{Relational Composition} task over unstructured text, showing that our invariant reasoning core can discover complex latent routing paths in natural language. We choose these tasks because their computational depth is controllable, enabling rigorous analysis of generalization behavior.
 
Our experiments reveal a consistent pattern that we term \emph{the computational frontier}: a diagonal boundary in the accuracy heatmap (thinking steps $\times$ task complexity). Once the number of thinking steps reaches the task's complexity, accuracy becomes near-perfect on the two structured tasks and reaches a lower plateau on unstructured text; the rise is abrupt from chance on graph reachability and gradual on the other two. We relate these different generalization profiles to the different inductive biases of their respective perception interfaces.
 
Our contributions are summarized as follows:
\begin{itemize}
    \item We study vertical CoT, which adds test-time reasoning compute by repeating a shared-weight block instead of generating tokens, and instantiate it as a reasoning core of under 1M parameters, trained via silent thinking, LayerScale initialization, and identity-biased gating.
    \item Our test-time depth recurrence extrapolates to problem depths and step counts beyond the training setup. On the graph task, it generalizes where fixed-depth Transformers barely extrapolate, and on the two sequence tasks, it comes within two points of fixed-depth models that use $4$--$6.4\times$ the parameters.
    \item We identify and analyze the computational frontier phenomenon and show that generalization behavior differs qualitatively across task-specific perception interfaces.
    \item We show that intermediate per-step supervision, a standard recipe for training deep iterative models, \emph{harms} OOD generalization on the graph task, whose adjacency mask lets us verify how far information has actually propagated.
    \item We show that the recurrent core holds peak inference memory flat and latency linear in the reasoning-step count, in contrast to the growing key--value cache and $\Theta(TN)$ critical path of horizontal CoT.
\end{itemize}

\section{Related Work}
\label{sec:related}

\subsection{Chain-of-Thought and Test-Time Compute}

Chain-of-Thought prompting~\citep{wei2022chain} and its variants~\citep{kojima2022large, wang2022self} have become the dominant paradigm for LLM reasoning. Recent work on test-time computing~\citep{snell2024scaling} demonstrates that allowing models to perform more inference-time computation improves performance. However, all these approaches operate via horizontal token generation, consuming a context window proportional to the reasoning depth. Our work explores an orthogonal axis, vertical depth recurrence in latent space, which achieves test-time compute scaling without token overhead.
 
\subsection{Pause Tokens and Latent Reasoning}

Researchers proposed appending learnable ``pause tokens'' to the input~\citep{goyal2023think}, giving the Transformer additional forward-pass computation before producing output; Coconut~\citep{hao2024training} takes this further by feeding the last hidden state back as the next input embedding, keeping intermediate ``thought'' steps continuous rather than discretized into vocabulary tokens. Both share our motivation of extra inference-time computation, and Coconut further shares our goal of computing in continuous space. However, both remain fundamentally \emph{horizontal}. Our approach instead increases depth by repeatedly applying the same block without growing the sequence. Unlike traditional RNNs that compress the sequence into a single bottleneck vector $h \in \mathbb{R}^d$, we maintain a full-sequence-length state matrix $H \in \mathbb{R}^{L \times d}$ at every step, preserving rich spatial interactions.
 
\subsection{Universal Transformers}

The Universal Transformer (UT)~\citep{dehghani2018universal} introduced weight sharing across layers and used Adaptive Computation Time (ACT)~\citep{graves2016adaptive} for dynamic stopping. Our work builds on this foundation but differs in three critical aspects. 
First, LayerScale~\citep{touvron2021going} protects latent representations during early training. 
Second, an identity-biased gated recurrence with negative bias initialization improves gradient propagation across many recurrence steps (Section~\ref{sec:gate-gradient}). Third, rather than ACT's complex token-level halting probabilities, we treat the recurrence step count $T$ as an externally specified budget that decouples computational depth from optimization, supporting flexible test-time compute scaling without optimization overhead. This choice is not merely a simplification: we found that an ACT Universal Transformer learns a halting budget that barely grows with the depth OOD instances require, falling to chance on graph reachability from 8 hops onward (Sections~\ref{sec:baselines} and~\ref{sec:limitations}).
 
\subsection{Weight Sharing and Fixed-Point Views of Depth}

Parameter sharing across layers has also been explored outside the recurrent-reasoning setting. ALBERT~\citep{lan2020albert} shares parameters across layers as a model-compression technique. However, it applies the same fixed number of layers to every input, with no mechanism to vary that count at test time. A complementary perspective comes from Deep Equilibrium Models (DEQ)~\citep{bai2019deep}, which reinterpret weight-sharing recurrence as a fixed point solved via root-finding rather than unrolling. We retain the intermediate iterates, which lets us control how the number of iterations $T$ trades off against accuracy.

\subsection{Depth and Expressiveness in Transformers}

Theoretical works have established that Transformer depth is critical in expressiveness: fixed-depth Transformers are limited to $\mathsf{TC}^0$ circuit complexity~\citep{merrill2023expresssive}, which excludes inherently sequential computations. Researchers demonstrated that depth-efficient Transformers cannot solve certain compositional tasks without sufficient layers~\citep{feng2023towards}. By making depth a dynamic variable through recurrence, our architecture bypasses the $\mathsf{TC}^0$ limitation: the unrolled reasoning steps can scale with the input's intrinsic sequential complexity, enabling the model to solve inherently sequential tasks, such as multi-hop routing and nested logic, that fixed-depth networks struggle with.

\subsection{Summary Comparison}

Table~\ref{tab:relwork} compares our approach with the closest families of test-time-compute methods along the axes discussed above.

\newcolumntype{Y}{>{\raggedright\arraybackslash}X}
\begin{table*}[tb]
    \centering
    \caption{Comparison of approaches for scaling test-time computation.}
    \label{tab:relwork}
    \footnotesize
    \setlength{\tabcolsep}{5pt}
    \begin{tabularx}{\linewidth}{@{}p{65pt}p{57pt}p{28pt}p{106pt}Yp{112pt}@{}}
        \toprule
        \textbf{Method} & \textbf{Compute Axis} & \textbf{Weight Sharing} & \textbf{Supervision} & \textbf{Context Cost} & \textbf{Stability Mechanism} \\
        \midrule
        Horizontal CoT~\citep{wei2022chain} & Token generation & No & Per-token / outcome & Grows with reasoning length & N/A (fixed depth per token) \\
        Pause Tokens~\citep{goyal2023think} / Coconut~\citep{hao2024training} & Token generation (latent) & No & Per-token / outcome & Grows with \# pause tokens & N/A (fixed depth per token) \\
        Universal Transformer~\citep{dehghani2018universal} & Layer recurrence & Yes & Halting-weighted average of all steps + ponder cost~\citep{graves2016adaptive} & None & ACT halting regularization \\
        Ours & Layer recurrence & Yes & Final output & None & LayerScale + identity-biased gate \\
        \bottomrule
    \end{tabularx}
\end{table*}

\section{Method}
\label{sec:method}

Our architecture consists of three components (Figure~\ref{fig:architecture}): a \emph{task-invariant reasoning core} that iteratively refines a hidden representation through shared-weight recurrence, a \emph{task-specific perception interface} that encodes the raw input into this representation's initial state, and a \emph{task-specific readout head} that extracts the final prediction after $T$ recurrence steps ($T$ can be flexibly scaled at inference time).

\begin{figure}[t]
    \centering
    \begin{tikzpicture}[
        node distance=0.58cm and 0.53cm,
        every node/.style={font=\footnotesize},
        block/.style={draw, rounded corners, align=center, minimum height=0.85cm, inner sep=3pt},
        >=Latex
    ]
        \node[block, fill=blue!8, minimum width=1.9cm] (perc) {Perception\\Interface};
        \node[block, fill=orange!12, minimum width=2.3cm, right=of perc] (core) {Reasoning Core\\$f_\theta$ (shared)};
        \node[block, fill=blue!8, minimum width=1.7cm, right=of core] (read) {Readout\\Head};

        \node[left=0.5cm of perc] (x) {$x$};
        \node[right=0.5cm of read] (y) {$\hat{y}$};

        \draw[->] (x) -- (perc);
        \draw[->] (perc) -- node[above, font=\scriptsize]{$H^{(0)}$} (core);
        \draw[->] (core) -- node[above, font=\scriptsize]{$H^{(T)}$} (read);
        \draw[->] (read) -- (y);
        \draw[->] (core) edge[loop above, looseness=6, min distance=12mm] node[above, font=\scriptsize]{$\times\, T,\; +\, e_t$} (core);
    \end{tikzpicture}
    \caption{Overview of the depth-recurrent architecture. A task-specific perception interface encodes the input into $H^{(0)}$; the task-invariant core $f_\theta$ refines this state recurrently for $T$ steps using a \emph{single} set of shared weights, conditioned at each step on a depth embedding $e_t$; a task-specific readout head decodes $H^{(T)}$ into the prediction $\hat{y}$. Unlike horizontal CoT, $T$ can be scaled at inference time without adding parameters or lengthening the input/output sequence.}
    \label{fig:architecture}
\end{figure}

\subsection{The Invariant Reasoning Core}
\label{sec:core}
The reasoning core is a single Transformer block $f_\theta$ with shared weights, applied recurrently for $T$ steps. Let $L$ be the sequence length and $d$ be the hidden dimension; the core operates on a hidden state $H^{(t)} \in \mathbb{R}^{L \times d}$, where the initial state $H^{(0)}$ is the output of the task-specific perception interface (Section~\ref{sec:perception}). At each step $t \in \{1, \dots, T\}$, the candidate hidden state $\tilde{H}^{(t)} \in \mathbb{R}^{L \times d}$ is computed as:
\begin{equation}
    \tilde{H}^{(t)} = f_\theta(H^{(t-1)}, t),
\end{equation}
Below we detail its architecture, together with three mechanisms that help it unroll stably for 20+ steps.

\paragraph{Architecture of $f_\theta$}
The block is a single Pre-LayerNorm Transformer encoder block~\citep{xiong2020layer}. At recurrence step $t$, it receives the depth-conditioned input $\hat{H}^{(t)} = H^{(t-1)} + e_t$, where $e_t$ is the depth embedding described below. The block applies two sub-layers in sequence.

\emph{Sub-layer 1: Multi-Head Self-Attention (MHSA).}
\begin{equation}
    H' = \hat{H}^{(t)} + \gamma_{\mathrm{attn}} \odot
         \mathrm{MHSA}\!\left(\mathrm{LN}\!\left(\hat{H}^{(t)}\right),\, M\right),
\end{equation}
where $\mathrm{LN}(\cdot)$ is Layer Normalization, $M$ is the task-specific attention mask (the adjacency mask for the topological interface, or no mask for the other two; see Section~\ref{sec:perception}), and $\gamma_{\mathrm{attn}} \in \mathbb{R}^d$ is a LayerScale vector. The attention mechanism computes $h$ parallel heads. For each head $i$:
\begin{align}
    \mathrm{head}_i &= \mathrm{softmax}\!\left(
        \frac{Q_i K_i^\top}{\sqrt{d_k}} + M
    \right) V_i, \\
    Q_i = \hat{H}^{(t)} W_i^Q,&\quad
    K_i = \hat{H}^{(t)} W_i^K,\quad
    V_i = \hat{H}^{(t)} W_i^V,
\end{align}
with $d_k = d / h$. For the hierarchical and unstructured interfaces, Rotary Position Embeddings (RoPE)~\citep{su2024roformer} are applied to the query and key projections before the dot product, so that relative position is encoded (see Section~\ref{sec:perception}).

\emph{Sub-layer 2: Feed-Forward Network (FFN).}
\begin{align}
    H'' &= H' + \gamma_{\mathrm{ffn}} \odot \mathrm{FFN}\!\left(\mathrm{LN}(H')\right), \\
    \mathrm{FFN}(x) &= W_2 \cdot \mathrm{GELU}(W_1 x + b_1) + b_2,
\end{align}
where $W_1 \in \mathbb{R}^{d_{\mathrm{ff}} \times d}$, $W_2 \in \mathbb{R}^{d \times d_{\mathrm{ff}}}$, and $d_{\mathrm{ff}}$ is the feedforward dimension. The output of $f_\theta$ is $\tilde{H}^{(t)} = H''$, which is subsequently combined with $H^{(t-1)}$ through the identity-biased gate (below) to produce $H^{(t)}$.

For the hierarchical and unstructured perception interfaces (Section~\ref{sec:perception}), we apply LayerScale~\citep{touvron2021going} after each sub-layer, with $\gamma_{\mathrm{attn}}, \gamma_{\mathrm{ffn}} \in \mathbb{R}^d$ initialized to $10^{-4}$ and learned jointly with all other parameters. For the topological interface, where the hard adjacency mask strongly constrains the computation, we omit LayerScale and instead use the standard PyTorch \texttt{TransformerEncoderLayer} to replace the custom RoPE-integrated attention described above. Table~\ref{tab:hparams} in Section~\ref{sec:experiments} summarizes the architectural hyperparameters.

\paragraph{Silent Thinking Objective}
A fundamental design choice is how to supervise the recurrent steps. Instead of forcing intermediate answers or applying per-step auxiliary losses, we compute the cross-entropy (CE) loss $\text{CE}(\hat{y},\, y)$ \emph{only} on the final thinking step $T$.

During training, $T$ is sampled uniformly from a task-specific range (e.g., $T \sim \mathcal{U}(1, 12)$). This randomized depth acts as a regularizer, encouraging the network to produce stable outputs across a range of computation depths, while the silent thinking objective forces the model to develop its own genuine multi-step reasoning paths. Algorithm~\ref{alg:training} summarizes one training step. At inference time, $T$ is set to any desired budget up to $T_{\max}$, independent of the range used during training.

\begin{algorithm}[t]
\caption{Training Step with Silent Thinking}
\label{alg:training}
\begin{algorithmic}[1]
\Require input $x$, label $y$, perception interface $\mathrm{Perc}$, readout head $\mathrm{Read}$, reasoning block $f_\theta$, training step range $[T_{\min}, T_{\max}^{\mathrm{tr}}]$
\State $H^{(0)} \gets \mathrm{Perc}(x)$
\State Sample $T \sim \mathcal{U}\{T_{\min}, \dots, T_{\max}^{\mathrm{tr}}\}$
\For{$t = 1, \dots, T$}
    \State $\hat{H}^{(t)} \gets H^{(t-1)} + e_t$ \Comment{depth embedding}
    \State $\tilde{H}^{(t)} \gets f_\theta(\hat{H}^{(t)})$
    \State $z^{(t)} \gets \sigma\!\left([\tilde{H}^{(t)}; H^{(t-1)}] W_z + b_z\right)$
    \State $H^{(t)} \gets z^{(t)} \odot \tilde{H}^{(t)} + (1 - z^{(t)}) \odot H^{(t-1)}$
    \Comment{no loss is computed here}
\EndFor
\State $\hat{y} \gets \mathrm{Read}(H^{(T)})$
\State $\mathcal{L} \gets \mathrm{CE}(\hat{y}, y)$ \Comment{loss only at the final step $T$}
\State Backpropagate $\mathcal{L}$ through the unrolled computation graph and update $\theta$
\end{algorithmic}
\end{algorithm}

\paragraph{Pre-Layer Normalization and LayerScale}

For complex symbolic tasks requiring a high step count, the accumulated variance from repeated sub-layer transformations can destabilize early training. Beyond the Pre-LayerNorm~\citep{xiong2020layer} placement above, initializing the LayerScale vectors $\gamma_{\mathrm{attn}}, \gamma_{\mathrm{ffn}}$ near zero ($10^{-4}$) forces the early-training dynamics of $f_\theta$ to act almost perfectly as an identity mapping, protecting fragile boolean states from the destructive noise introduced by randomly initialized deep layers. As training progresses, the network selectively scales up $\gamma$ to activate its reasoning capacity.

\paragraph{Identity-Biased Gated Recurrence}

To backpropagate through many unrolled training steps without vanishing gradients, we combine the new candidate representation $\tilde{H}^{(t)}$ with the previous state $H^{(t-1)}$ through a learned GRU-like gate:
\begin{align}
    z^{(t)} &= \sigma\!\left([\tilde{H}^{(t)}; H^{(t-1)}] \cdot W_z + b_z\right) \\
    H^{(t)} &= z^{(t)} \odot \tilde{H}^{(t)} + (1 - z^{(t)}) \odot H^{(t-1)}
\end{align}

Here, $[\tilde{H}^{(t)}; H^{(t-1)}] \in \mathbb{R}^{L \times 2d}$ denotes the feature-dimension concatenation of the two states, and $W_z \in \mathbb{R}^{2d \times d}$ projects it to yield $z^{(t)} \in \mathbb{R}^{L \times d}$. The gate operates element-wise at each sequence position.

We initialize the gate bias $b_z = -2.0$. Because $\sigma(-2.0) \approx 0.12$, the model defaults to retaining 88\% of its previous state. This physically guarantees that the initial signal decays slowly, enabling stable signal propagation through deep unrolling. We formalize this intuition with a gradient analysis in Section~\ref{sec:gate-gradient}.

\paragraph{Depth Embeddings}

To prevent the shared-weight block $f_\theta$ from losing its place within the recurrent loop, we inject a learned depth embedding $e_t \in \mathbb{R}^d$ at each step $t \in \{1, \dots, T\}$ before the attention computation:
\begin{equation}
    \hat{H}^{(t)} = H^{(t-1)} + \mathbf{1}_L \cdot e_t^\top,
\end{equation}
where $\mathbf{1}_L \in \mathbb{R}^L$ is an all-one vector, so that $e_t$ is broadcast identically across all $L$ sequence positions. The embeddings $\{e_t\}_{t=1}^{T_{\max}}$ form a standard learned lookup table $E \in \mathbb{R}^{T_{\max} \times d}$, where $T_{\max}$ is set to cover the maximum number of steps used at both training and test time.

Without depth embeddings, the shared block $f_\theta$ has no signal indicating which recurrence step it is on, making it difficult to distinguish early exploratory steps from later refinement steps. Depth embeddings are orthogonal to the positional encodings used within the sequence (RoPE or none): the former encodes \emph{which recurrence step} the state is currently at, the latter encodes \emph{which sequence position} a token occupies, and both are applied simultaneously without conflict.

\subsection{Task-Specific Perception Interfaces}
\label{sec:perception}

Rather than hardcoding task-specific rules into the reasoning core, we design modular perception interfaces that inject varying degrees of structural inductive bias through the initial state $H^{(0)}$ and through the attention mask and positional encoding that the core applies at every step. The core's design is shared across tasks, but each task trains its own core; ``task-invariant'' refers to this shared recipe, not to a single trained model. We explore three distinct interface paradigms:

\paragraph{Topological Masking (Strict Structural Priors)}

To enforce physical constraints where computation must strictly follow defined topological pathways (e.g., graph routing), we introduce an adjacency mask. Given an adjacency matrix $A \in \{0,1\}^{L \times L}$ over the $L$ sequence positions, we add self-loops ($A_{ii} = 1$) and construct an additive mask matrix $M \in \mathbb{R}^{L \times L}$:
\begin{equation}
M_{ij} =
\begin{cases}
0, & \text{if } A_{ij} = 1 \\
-\infty, & \text{if } A_{ij} = 0
\end{cases}
\end{equation}
The attention is computed as $\text{softmax}(QK^T/\sqrt{d_k} + M)V$. The $-\infty$ term forces the attention probability to zero for non-adjacent nodes.

\paragraph{Relative Positional Encoding (Hierarchical Priors)}

For sequences where structural hierarchy depends on relative distances rather than strict adjacency (e.g., evaluating nested symbolic expressions), topological masking is overly restrictive. Instead, we utilize \emph{Rotary Position Embeddings (RoPE)}~\citep{su2024roformer}. By multiplying the query and key vectors with rotation matrices corresponding to their respective absolute positions $m$ and $n$ ($R_{\Theta, m}$ and $R_{\Theta, n}$), the inner product inherently encodes their relative sequence distance $(m - n)$:
\begin{equation}
\langle \tilde{q}_m, \tilde{k}_n \rangle = q^T R_{\Theta, m-n} k
\end{equation}

\paragraph{Standard Sequence Attention (Unstructured / Task-Agnostic Priors)}

To test the reasoning core's ability to autonomously discover latent routing paths in unstructured sequences (e.g., natural language facts), we remove task-specific structural biases. Although we retain standard Rotary Position Embeddings (RoPE) to let the perception interface process local word order, we completely shuffle the input facts. Consequently, unlike the previous experiment, where relative distance correlates with hierarchical depth, the 1D sequence distance in this bag-of-facts provides no meaningful structural hints about the underlying relational graph. The invariant reasoning core must discover the correct pointer-chasing routes entirely on its own.

\subsection{Task-Specific Readout Mechanisms}
\label{sec:readout}

After unrolling the latent reasoning for $T$ steps, a readout head decodes the final state $H^{(T)}$. We employ distinct readout mechanisms depending on the interface:
\begin{itemize}
    \item Pairwise Node Readout (for the topological interface): The $d$-dimensional representations corresponding to the specific source and target nodes are extracted, concatenated, and passed through an MLP.
    \item Global Sequence Readout (for the hierarchical and unstructured interfaces): The global representation of the \texttt{[CLS]} token at position 0 is extracted and passed through an MLP. Because neither interface exposes the query positions to the readout, the reasoning core must route the relevant information to the \texttt{[CLS]} token by the final step.
\end{itemize}

\section{Experiments} \label{sec:experiments}
We evaluate our architecture on three compositional reasoning domains with decreasing structural inductive biases: graph reachability, Boolean logic, and relational text. The maximum number of evaluated reasoning steps varies across these tasks, reflecting the natural upper bounds of their respective structural complexities.

Table~\ref{tab:hparams} summarizes the architectural hyperparameters used for each experiment. Graph reachability's small node count requires less representational capacity than the richer symbolic and lexical vocabularies of the logic and family tasks, so we use a narrower $d$ for efficiency.

\begin{table}[tb]
    \centering
    \caption{Architectural hyperparameters for each experiment. Depth-embedding capacity is the number of depth embeddings $T_{\max}$, i.e., the largest step count the model can run; the maximum step count actually evaluated in Sections~\ref{sec:graph}--\ref{sec:text} is 20 (graph), 24 (nested expr.), and 20 (family).}
    \label{tab:hparams}
    \begin{tabular}{lccc}
        \toprule
        & \textbf{Graph} & \textbf{Nested Expr.} & \textbf{Family} \\
        \midrule
        $d$            & 128  & 256  & 256  \\
        $h$ (heads)    & 4    & 8    & 8    \\
        $d_\mathrm{ff}$& 256  & 1024 & 1024 \\
        RoPE           & \texttimes & $\checkmark$ & \checkmark \\
        LayerScale     & \texttimes & $\checkmark$ & \checkmark \\
        Gate bias $b_z$& $-2.0$ & $-2.0$ & $-2.0$ \\
        Train step range $T$ & $[5, 8]$ & $[4, 16]$ & $[1, 12]$ \\
        Depth-emb.\ capacity & 20 & 28 & 20 \\
        \bottomrule
    \end{tabular}
\end{table}

\subsection{Experiment I: Graph Reachability}
\label{sec:graph}

Given a directed graph $G = (V, E)$, this task determines whether there exists a directed path from node $s$ to $r$. The model is trained on instances requiring 1--5 hops with 5--8 thinking steps and is evaluated up to 12 hops to test OOD generalizability. Throughout this section, each heatmap marks both the depth and thinking-step training boundaries with dashed lines, with axes inside/outside those ranges denoting ID/OOD evaluation.

\begin{figure}[tb]
    \centering
    \includegraphics[width=.5\columnwidth]{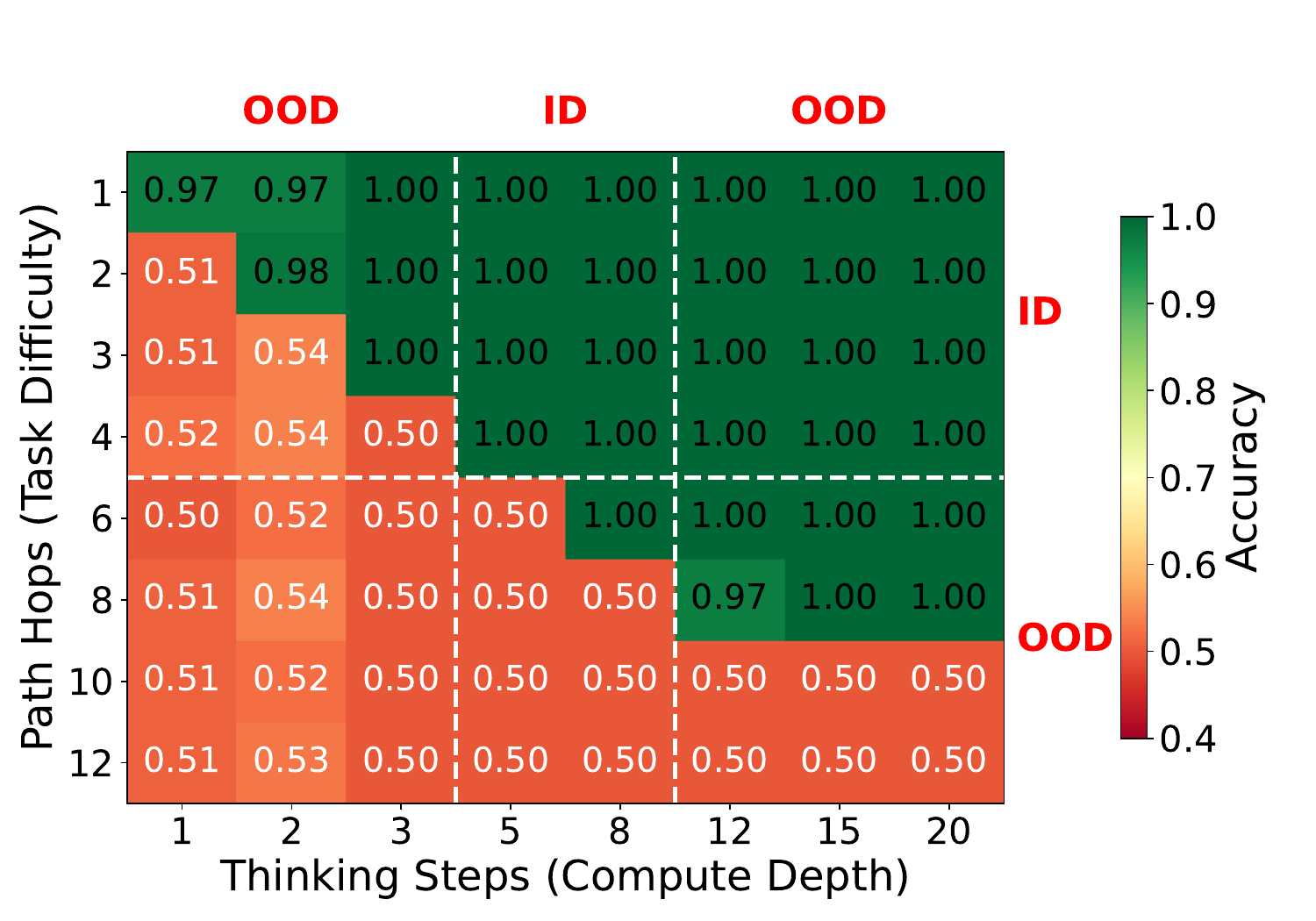}
    \caption{Accuracy heatmap for the graph reachability task. A sharp computational frontier exists along the diagonal, confirming the strict 1-hop $=$ 1-step physical constraint enforced by adjacency masking. The model ($d{=}128$) generalizes perfectly to 8 hops (OOD) with sufficient steps, but collapses abruptly to chance (${\sim}50\%$) at 10 hops. Dashed lines mark the training boundaries in both dimensions (1--5 hops; 5--8 thinking steps); cells outside these ranges are OOD. Up to 8 hops, accuracy is preserved wherever the thinking-step count meets or exceeds the hop count.}
\label{fig:graph-heatmap}
\end{figure}

As illustrated in Figure~\ref{fig:graph-heatmap}, we observe a sharp computational frontier. The accuracy jumps from chance to perfect/near-perfect: once given enough ($N+$) thinking steps, the model solves $N$-hop queries. One step fewer and accuracy drops to random. The model achieves 100\% OOD generalization up to 8 hops (1.6$\times$), but collapses abruptly at 10 hops. This boundary is set by capacity rather than by the mask: a wider model also solves 10-hop queries (Section~\ref{sec:ablation}). In the thinking-step dimension, the model also generalizes stably both below (1--3 thinking steps) and above (12--20 thinking steps) the training range, with the diagonal frontier shifting accordingly.

\paragraph{Why Is the Frontier Sharp?}
Consider the 5-node chain graph $s \to v_1 \to v_2 \to v_3 \to r$ (a 4-hop query). Under the adjacency mask $M$ (Section~\ref{sec:perception}), each application of $f_\theta$ moves information across at most one edge, so $r$'s hidden state reflects $s$'s existence only once $t \ge 4$; since the pairwise readout (Section~\ref{sec:readout}) queries $s$ and $r$ directly, it can only report ``reachable'' once this frontier physically arrives, which is exactly why accuracy collapses to chance one step below the required hop count. Once it arrives, the identity-biased gate (Section~\ref{sec:core}) favors retaining it, so extra steps rarely hurt.

\subsection{Experiment II: Nested Boolean Expression Evaluation}
\label{sec:logic}

Given a fragile nested boolean expression (e.g., \texttt{!((T\&F)|(!(T|F)))}), this task evaluates it to \texttt{True} or \texttt{False}. Training succeeds on nesting depths 1--8.

\begin{figure}[tb]
    \centering
    \includegraphics[width=.5\columnwidth]{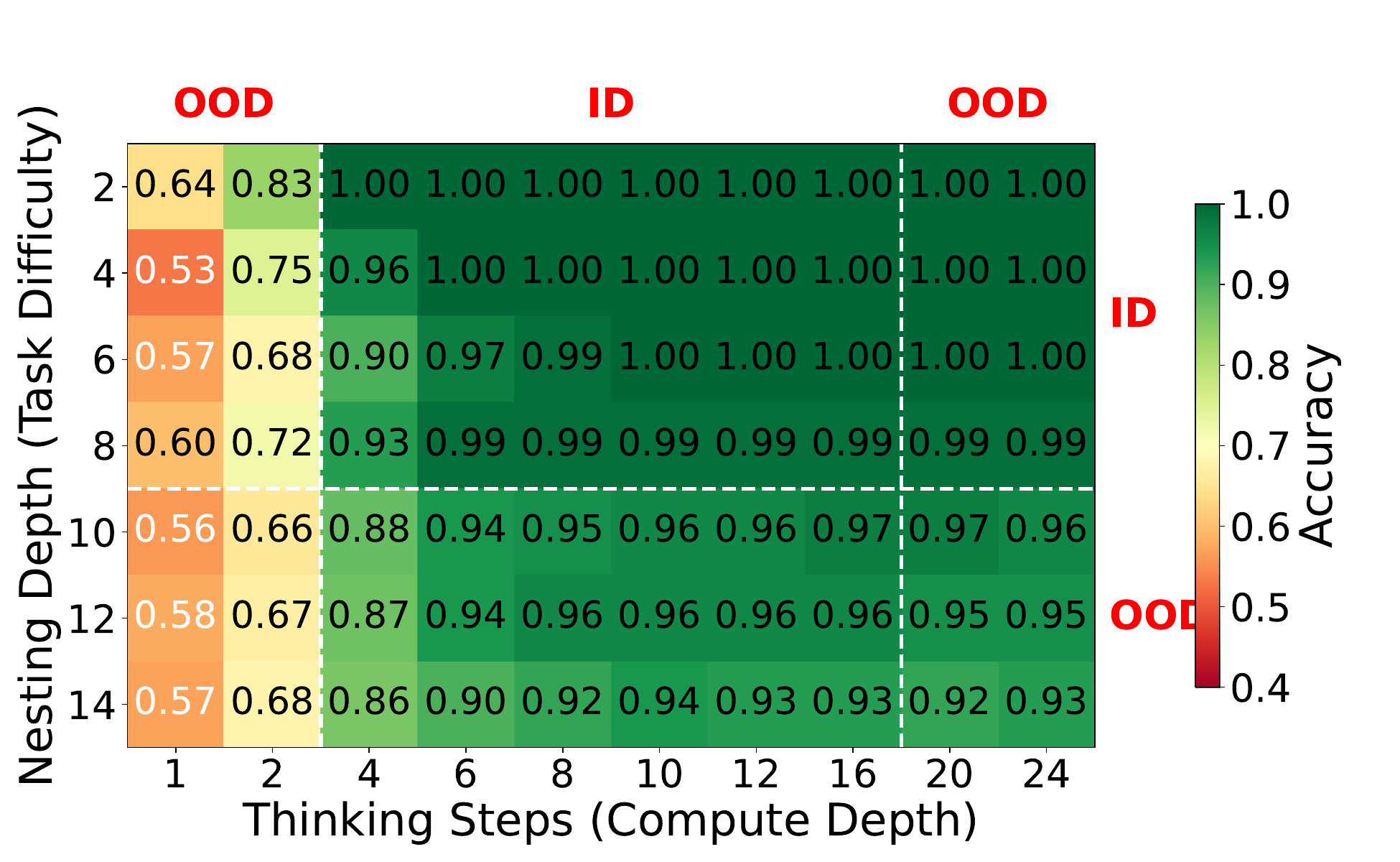}
    \caption{Accuracy heatmap for the nested boolean expression task. Unlike the graph task, the computational frontier is gradual and the model degrades gracefully on OOD samples, holding  $90\%+$ accuracy up to depth 14 when given sufficient thinking steps, and remaining stable even when unrolled for 24 thinking steps (well beyond the training range of 4--16 thinking steps). Dashed lines mark the training boundaries in both dimensions (1--8 depth; 4--16 thinking steps).}
\label{fig:logic-heatmap}
\end{figure}

As depicted in Figure~\ref{fig:logic-heatmap}, we observe a gradual computational frontier. The model degrades gracefully beyond the training distribution, holding $90\%+$ accuracy at depth 14 (1.75$\times$ OOD). Rather than jumping from chance at a fixed step count as in the graph task, accuracy increases gradually and monotonically with more thinking steps, and the model does not ``overthink'' or degrade with excessive computation up to 24 steps (OOD in the step dimension), consistent with the identity-biased gate favoring state retention (Section~\ref{sec:core}).

\paragraph{Why Is Degradation Graceful?}
Unlike the hard adjacency mask used for graph reachability, RoPE only \emph{biases} attention toward nearby positions rather than blocking distant ones: rotating $q$ and $k$ by relative angles makes their dot product decay, on average, as the relative distance grows~\citep{su2024roformer}, so even in a single step $f_\theta$ can already attend weakly to operands several nesting levels away. A model given fewer than 4 thinking steps is therefore not blind to the outer operations, unlike the graph model; it can partially propagate operand values across levels in one step, at the cost of interference between them. This is why accuracy degrades gradually instead of collapsing sharply below the required depth (Figure~\ref{fig:logic-heatmap}).

\subsection{Experiment III: Relational Composition over Text}
\label{sec:text}

To test the reasoning core's capability in a pure language modeling context, we evaluate it on a family-vocabulary relational composition task inspired by CLUTRR~\citep{sinha2019clutrr}, where the correct relation is defined by a signed-offset algebra over parent/child hops rather than literal family semantics. The input is a sequence of shuffled natural-language sentences that define relationships (e.g., \texttt{Alice is the parent of Bob}), padded with irrelevant distractor relations, followed by a query statement asserting a candidate relation between two entities (e.g., \texttt{Alice is the sibling of Eve}). The model determines whether the queried relation is correct.

Shuffling the fact sentences and padding them with irrelevant distractors already prevents shortcuts based on sentence position or sentence count. Na\"ively generating chains of relations (e.g., strictly parents or strictly children), however, still allows models to cheat by counting relation words, since the label would then be a monotonic function of chain length. To enforce true algorithmic deduction (pointer chasing and offset cancellation), we generate chains using an \emph{Apex Routing Strategy}: the path alternates a \texttt{parent}-hop phase (offset $+1$ each) and a \texttt{child}-hop phase (offset $-1$ each), so \texttt{Alice} $\to$ \texttt{Carol} $\to$ \texttt{Eve} composes to $(+1)+(-1)=0$, the offset assigned to \texttt{sibling}. However, since offset $= 2u - K$ for a chain of length $K$ with $u$ up-hops, the offset and $K$ always share the same parity, leaving a residual \emph{parity-detection} shortcut: a model could accept or reject a candidate purely from whether its parity matches the chain length, without recovering the exact value. We close this loophole with \emph{hard negatives} whose parity matches the correct relation (e.g., \texttt{grandparent} for \texttt{sibling}, both even), forcing genuine latent routing rather than surface pattern matching. Fixing this single-reversal shape across all instances also means that increasing the chain depth changes only the length of the signed-offset sum, not the topology the model must recognize, isolating this ability from any confound with path-shape complexity.

The model is trained on depths 2--5 with 1--12 thinking steps, and evaluated up to depth 9 and 20 thinking steps.

\begin{figure}[tb]
    \centering
    \includegraphics[width=.5\columnwidth]{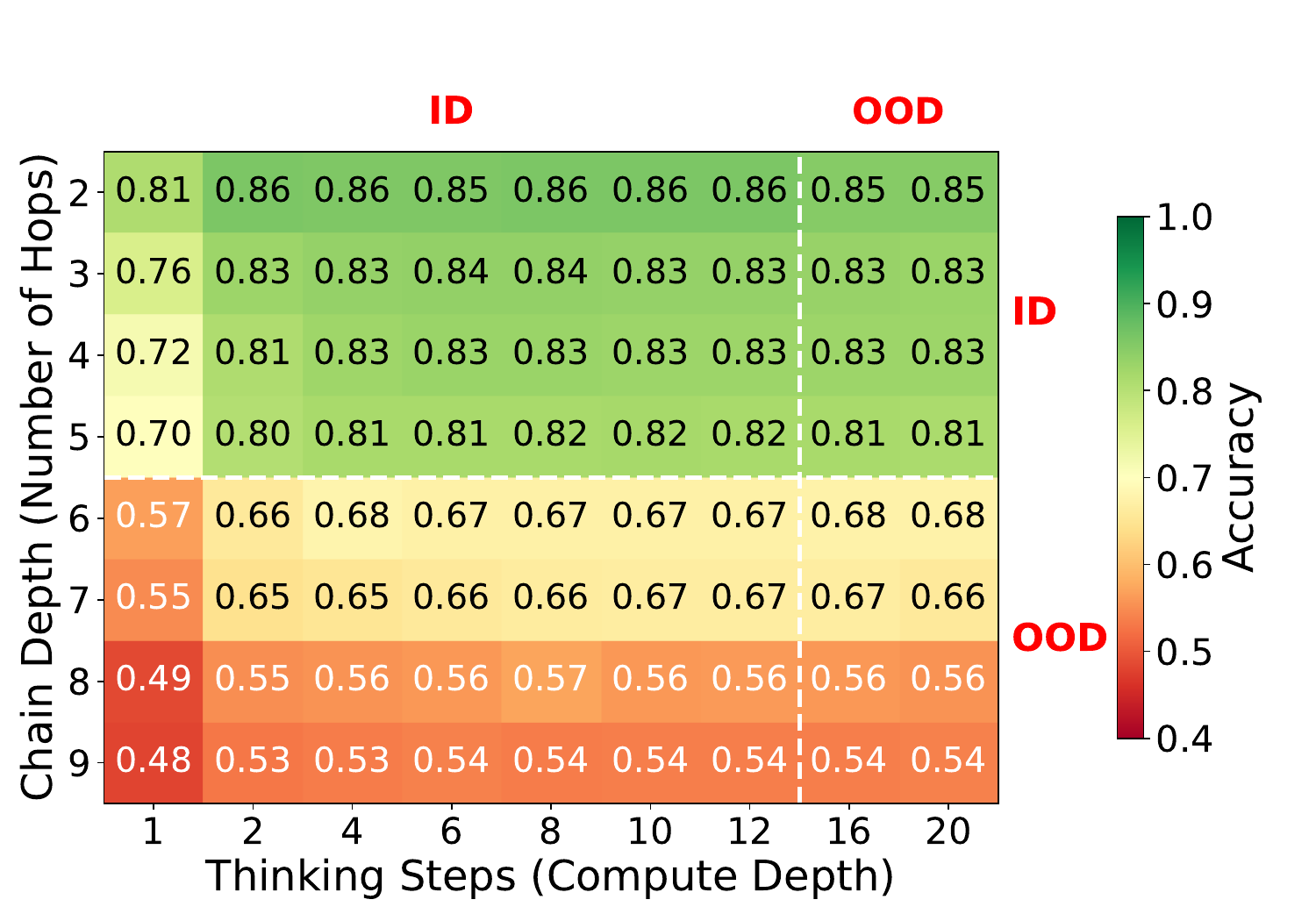}
    \caption{Accuracy heatmap for the relational composition task over unstructured text. Without task-aligned structural inductive biases (no adjacency masking; while RoPE is retained for local word order, the fully shuffled input facts ensure that 1D relative positions carry no meaningful structural signal), the task exhibits a monotonic increase in difficulty. The invariant reasoning core discovers latent pointer-chasing routes, staying above chance at depths 6 and 7 with more thinking steps, though accuracy falls toward chance by depth 9. Dashed lines mark the training boundaries in both dimensions.}
\label{fig:text-heatmap}
\end{figure}

As shown in Figure~\ref{fig:text-heatmap}, the results reveal three crucial insights. First, monotonic difficulty: the accuracy decreases overall as the chain depth increases. Because all parity shortcuts are removed, the model must perform genuine latent routing, which naturally becomes harder for longer chains. Second, the computational frontier: for any given depth, increasing the number of thinking steps improves accuracy up to a plateau. Third, partial OOD generalization: despite lacking task-aligned structural hints (no graph mask, and 1D relative positions provide no shortcuts for the fully shuffled sentences), the invariant core stays above chance at depths 6 and 7 ($66\%+$ once given sufficient steps), and unrolling past the trained range (16--20 steps) neither helps nor hurts; accuracy holds flat. It, nonetheless, approaches chance by depth 9.

\paragraph{Why Does Accuracy Decrease Monotonically? A Worked Example}
Unlike graph reachability (hard adjacency mask) and nested boolean logic (RoPE's soft bias still lets one step partially propagate across levels), the family task offers no such safety net: each additional hop is another opportunity to route to the wrong entity, with no guarantee a misrouted step is later corrected. If each hop is resolved correctly with probability $p<1$ and errors are not selectively repaired, the probability of resolving a depth-5 chain falls off roughly as $p^5$---a compounding cost rather than a threshold effect. This is why Figure~\ref{fig:text-heatmap} decreases smoothly, with neither a sharp cutoff (Figure~\ref{fig:graph-heatmap}) nor a high-accuracy floor (Figure~\ref{fig:logic-heatmap}). Additional steps still raise $p$ substantially per hop---accuracy at depth 5 improves from $69.8\%$ at 1 step to $81.7\%$ at 12 steps---but even at 20 steps it plateaus below the other two tasks, since the model must still route through unstructured, shuffled text with no positional shortcut to lean on.

\subsection{Comparison to Fixed-Depth and Weight-Sharing Baselines}
\label{sec:baselines}

\begin{table*}[tb]
    \centering
    \caption{In-distribution (ID) and out-of-distribution (OOD) accuracy (\%) per task, with OOD measured at sufficient thinking steps; \textbf{P} is parameters in millions, counting the perception interface and readout head as well as the reasoning core; the reasoning core alone stays under 1M in every task ($0.17$--$0.93$M). Recurrent models are unrolled to $T_{\max}$ at inference. Every ID and OOD number is a 3-seed mean; in the OOD column, $\pm$ is one standard deviation, omitted when it rounds to zero. The last row ablates our stability recipe (no LayerScale, no gate); the graph core has no LayerScale, so there only the gate is removed. The Fixed-depth GNN can only be applied to the graph task; the sequence tasks have no graph to message-pass over.}
    \label{tab:baselines}
    \small
    \setlength{\tabcolsep}{4pt}
    \begin{tabularx}{\textwidth}{c>{\raggedright\arraybackslash}Xccccccccc}
        \toprule
        & & \multicolumn{3}{c}{\textbf{Graph}} & \multicolumn{3}{c}{\textbf{Nested Expr.}} & \multicolumn{3}{c}{\textbf{Family}} \\
        \cmidrule(lr){3-5}\cmidrule(lr){6-8}\cmidrule(lr){9-11}
        & \textbf{Model} & \textbf{P} & \textbf{ID} & \textbf{OOD} & \textbf{P} & \textbf{ID} & \textbf{OOD} & \textbf{P} & \textbf{ID} & \textbf{OOD} \\
        \midrule
        \multirow{2}{2.1cm}{\emph{Fixed depth}}
        & Fixed-depth GNN         & $0.5$ & $100$ & $51$ & --- & --- & --- & --- & --- & --- \\
        & Fixed-depth Transformer & $0.8$ & $100$ & $57{\pm}1$ & $6.4$ & $100$ & $97$ & $4.0$ & $81$ & $58{\pm}2$ \\
        \midrule
        \multirow{3}{2.1cm}{\emph{Weight-sharing / recurrent}}
        & Universal Transformer (ACT)~\citep{dehghani2018universal,graves2016adaptive} & $0.2$ & $100$ & $62$ & $0.9$ & $97$ & $91{\pm}2$ & $0.9$ & $82$ & $57{\pm}2$ \\ \addlinespace[3pt]
        & Depth-Recurrent (ours) & $0.2$ & $100$ & $75{\pm}0.2$ & $1.0$ & $100$ & $95{\pm}1$ & $1.0$ & $83$ & $60{\pm}1$ \\ \addlinespace[3pt]
        & Depth-Recurrent w/o LayerScale, w/o gate (ablation) & $0.2$ & $100$ & $71{\pm}6$ & $0.9$ & $100$ & $97{\pm}1$ & $0.9$ & $82$ & $62{\pm}2$ \\
        \bottomrule
    \end{tabularx}
\end{table*}

Table~\ref{tab:baselines} places our instantiation against fixed-depth and weight-sharing alternatives. Four things stand out. First, extrapolation on structured problems requires test-time depth: on graph reachability, the fixed-depth GNN sits at chance beyond the training range (1--5 hops), and the fixed-depth Transformer extrapolates just one hop further ($77\%$ at 6 hops) before falling to chance at 8 hops and beyond, whereas our model solves up to 8 hops. Second, ablating our stability recipe (last row: no LayerScale, no gate) leaves ID and mean OOD accuracy essentially unchanged, even slightly higher on the two sequence tasks, but destabilizes training on graph: the ablated model's graph OOD swings from $62$ to $76$ across seeds where ours varies by $0.2$. Since the graph core uses no LayerScale, this effect comes from the gate alone; on the two sequence tasks, the spread across seeds is similar with or without LayerScale and the gate ($\pm 1$--$2$ points). Third, on the sequence tasks, our model and its ablation are at most two points below a fixed-depth Transformer that uses $4$--$6.4\times$ our parameters. Fourth, the genuine Universal Transformer---weight tying plus ACT adaptive halting and a ponder cost---is the weakest recurrent variant: on graph, it still solves 6-hop queries ($96$--$99.6\%$) but falls to chance from 8 hops onward (exactly $50$ on hops $8$--$12$; OOD mean $62$). Its learned halting stays shallow: the ponder count at the source and target nodes rises only from ${\approx}3$ at 1 hop to ${\approx}4.5$ at 8--12 hops, while the ponder penalty rewards early stopping. This per-node count is only a proxy for propagation depth, since halting freezes a node's own updates while nodes still running can keep relaying information, which is consistent with 6-hop queries being solved at ${\approx}4.3$ ponder steps. The degradation is milder on the sequence tasks ($91$ on logic, $57$ on family). An externally specified step count $T$ avoids this coupling between halting and optimization (Section~\ref{sec:limitations}).

\subsection{Silent Thinking vs. Intermediate Supervision}
\label{sec:ablation}

A natural hypothesis for improving algorithmic reasoning is to apply \emph{intermediate supervision}~\citep{lee2015deeply, newell2016stacked, wu2022associated}: computing and averaging the loss at every thinking step.  We compare our silent-thinking objective against an intermediate-supervision baseline on the graph reachability task, which is ideal for diagnosis because the adjacency mask provides a physically verifiable ground truth: a model taking $k$ steps can only aggregate information from $k$ hops away.

We interpolate between two objectives using parameter $\alpha$.
\begin{equation}
\mathcal{L}(\alpha) = (1-\alpha)\,\text{CE}\big(\hat{y}^{(T)}, y\big) + \frac{\alpha}{T}\sum_{t=1}^{T}\text{CE}\big(\hat{y}^{(t)}, y\big),
\label{eq:blended-loss}
\end{equation}
where $\hat{y}^{(t)}$ is the readout applied to $H^{(t)}$, so $\alpha = 0$ recovers silent thinking; $\alpha = 1$ is per-step supervision. We deliberately widen the dimension to $d=256$ because with smaller capacity ($d=128$), \emph{neither} objective propagates past $8$ hops.

\begin{table*}[tb]
    \centering
    \caption{Supervision strategies on graph reachability. Step-1 Acc.\ 12-hop measures accuracy after a single thinking step on 12-hop queries, where the adjacency mask bounds true accuracy at $\approx 50\%$; any excess indicates a statistical shortcut. Suff.-step OOD averages the best accuracy over the OOD hop counts $\{6, 8, 10, 12\}$ once given enough thinking steps, including 12-hop, which stays at chance for every configuration; 10-hop is the deepest path any configuration solves.}
    \label{tab:ablation-inter}
    \begin{tabular}{lcccc}
        \toprule
        \textbf{Objective} & \textbf{Step-1 Acc.\ 12-hop} $\downarrow$ & \textbf{10-hop OOD} $\uparrow$ & \textbf{Suff.-step OOD} $\uparrow$ & \textbf{Train acc.\ (step-mean)} \\
        \midrule
        Silent thinking ($\alpha = 0$)  & $50.0\%$ & \textbf{84\%} & \textbf{83.6\%} & $87.6\%$ \\
        Light per-step ($\alpha = 0.1$) & $69.4\%$ & $50\%$        & $75.0\%$        & $94.1\%$ \\
        Full per-step ($\alpha = 1$)    & $65.4\%$ & $51\%$        & $74.8\%$        & 96.9\% \\
        \bottomrule
    \end{tabular}
\end{table*}

Table~\ref{tab:ablation-inter} is the average over 5 runs. The intermediate supervision exhibits an apparent anomaly: it achieves roughly $65$--$70\%$ accuracy on 12-hop paths after only a \emph{single} thinking step. Under strict topological masking, a 1-step model has no information about nodes 12 hops away. This validates that the model abandons genuine message-passing and likely learns statistical heuristics, such as estimating reachability from graph density or the source node's degree, for early-step optimization.

Varying $\alpha$ produces a threshold effect, not a graded trade-off. Silent thinking is the only setting that turns the extra width into deeper propagation: it solves 10-hop paths ($84\%$ at $d = 256$; $88$--$100\%$ for all 5 seeds at $d = 384$), and its sufficient-step OOD accuracy grows with capacity ($75\% \to 83.6\% \to 87\%$ at $d = 128/256/384$). This property disappears even when $\alpha = 0.1$: 10-hop accuracy falls to chance for every seed. Full per-step supervision ($\alpha = 1$) shows the same ceiling: OOD accuracy stays flat across the same widths ($75.1\% \to 74.8\% \to 75.1\%$) even as its mean per-step training accuracy climbs monotonically ($94.7\% \to 96.9\% \to 98.3\%$).

We attribute this to a \emph{bandwidth occupation} failure mode. When penalized at Step 1, the model faces a choice: adopt an ``honest strategy'' (accepting 50\% early accuracy to learn deep propagation) or a ``shortcut strategy'' (learning shallow heuristics for immediate reward). Intermediate supervision makes the shortcut irresistible. Once committed to these heuristics, the model loses the incentive to develop the true sequential algorithm and fails on deep OOD paths even when granted many steps at test time: as shown above, added width lets it fit the intermediate readouts more tightly without translating into deeper propagation.

\section{Analysis and Discussion}
\label{sec:discussion}

\subsection{Precise vs.\ Robust Generalization}
 
The most revealing finding is the difference in generalization behavior across the three tasks under the same reasoning-core recipe. Graph reachability (adjacency masking) is precise but brittle: the strong structural prior requires one thinking step per hop; OOD generalization is perfect up to 1.6$\times$ and then collapses, though added width extends this reach (Section~\ref{sec:ablation}). It is also the one task where a fixed-depth Transformer, even with several times more parameters, extrapolates only one hop before falling to chance (Table~\ref{tab:baselines}). Boolean logic (RoPE) is approximate but robust: the relative-positioning bias allows flexible propagation patterns, giving graceful degradation ($>$ 90\% accuracy at 1.75$\times$ OOD). Unstructured text is the hardest to optimize but degrades slowly rather than collapsing, reaching modest but above-chance OOD accuracy by chasing pointers purely through latent representations.
 
This suggests a trade-off: stronger inductive biases come with precise but brittle extrapolation, and weaker ones with less precise but more robust extrapolation.

\subsection{Depth-Embedding Extrapolation Is Not Confounded}
\label{sec:depth-emb-check}

Because the step count $T$ is sampled from a bounded range during training, thinking steps beyond that range use depth embeddings $e_t$ that never receive a gradient update and remain at their random initialization, a potential confound for the OOD claims in Sections~\ref{sec:graph} to \ref{sec:text}: apparent stability there could reflect these embeddings being effectively ignored rather than genuine architectural generalization. We tested this by substituting the untrained embeddings at inference with (i) zeros, (ii) the last in-distribution embedding (called clamping), and (iii) three independent random reinitializations, then re-evaluating each trained checkpoint on the same beyond-range grid. We found that the four variants (baseline/zero/clamp/reinit) agree closely on all three tasks in this region---accuracy differs by at most $0.5$ percentage points across variants (e.g., $0.744$ vs.\ $0.739$ for graph) and by $\le 0.2$ points across reinitialization seeds. The gate value $z^{(t)}$ also shows no discontinuity at the train/beyond-range boundary on any task. We conclude that the reported beyond-range stability is a genuine property of the identity-biased recurrence rather than an artifact of the specific untrained embedding values. As a byproduct, this also confirms that the graph task's collapse beyond 8 hops (Section~\ref{sec:graph}, at $d = 128$) is not rescued by clamping to a trained embedding.

\subsection{Why Negative Gate Bias Prevents Vanishing Gradients}
\label{sec:gate-gradient}

The intuitive justification for $b_z = -2.0$ (Section~\ref{sec:core}) is that it makes the model retain 88\% of its previous state by default; here we make precise why this stabilizes gradient flow across the up to $16$ steps unrolled in training. Differentiating the gated update $H^{(t)} = z^{(t)} \odot \tilde{H}^{(t)} + (1-z^{(t)}) \odot H^{(t-1)}$ and, following standard gated-recurrence gradient analyses~\citep{gers2000learning}, treating $z^{(t)}$ as approximately constant with respect to $H^{(t-1)}$ over a single step, the Jacobian decomposes as
\begin{equation}
    \frac{\partial H^{(t)}}{\partial H^{(t-1)}} \approx \mathrm{diag}\!\left(1 - z^{(t)}\right) + \mathrm{diag}\!\left(z^{(t)}\right) \frac{\partial \tilde{H}^{(t)}}{\partial H^{(t-1)}}.
\end{equation}
The first term bypasses $f_\theta$; the second routes through the reasoning block and is damped by LayerScale, whose $\gamma_{\mathrm{attn}}, \gamma_{\mathrm{ffn}}$ are initialized near zero ($10^{-4}$, Section~\ref{sec:core}) so that this branch contributes negligible signal early in training. Chaining the identity-path term across all $T$ steps, the gradient of the silent-thinking loss with respect to the initial state becomes
\begin{equation}
    \frac{\partial \mathcal{L}}{\partial H^{(0)}} \; = \; \frac{\partial \mathcal{L}}{\partial H^{(T)}} \prod_{t=1}^{T} \mathrm{diag}\!\left(1 - z^{(t)}\right) \; + \; R,
\end{equation}
where $R$ collects the remaining chain-rule terms that route through $f_\theta$ at least once at some step.
At initialization $W_z \approx 0$, so $z^{(t)} \approx \sigma(b_z)$ is approximately constant across steps and positions. With $b_z = -2.0$, $(1-z^{(t)}) \approx 0.88$, so this path attenuates the gradient by only $0.88^{T}$ (e.g., $0.88^{16} \approx 0.13$), a modest decay that keeps the gradient in a numerically healthy range, in contrast to the symmetric initialization $b_z = 0$, where $(1-z^{(t)}) \approx 0.5$ and the same calculation gives $0.5^{16} \approx 1.5 \times 10^{-5}$, nearly four orders of magnitude smaller in the vanishing-gradient regime. Because this identity path does not route through the LayerScale-suppressed branch, it remains the dominant channel carrying gradient signal back to $H^{(0)}$ and the perception interface. LSTMs justify large forget-gate bias initialization similarly~\citep{gers2000learning}: a near-linear self-connection lets the gradient signal pass through many steps largely unchanged. We adapt that argument to a Transformer-based recurrent core. As training progresses and $W_z$ is learned, $z^{(t)}$ departs from this constant profile to encode instance-specific update decisions.

\subsection{Computational and Memory Complexity}
\label{sec:complexity}

The following comparison focuses on the cost of \emph{thinking}, not producing the final answer, which is a single token in both paradigms. Throughout, $T$ denotes the depth of this reasoning process: $T$ recurrence steps for our vertical CoT, or $T$ reasoning tokens generated before the answer for horizontal CoT.

Because $f_\theta$ is a single block applied recurrently, its parameter count, $\Theta(d^2 + d\,d_{\mathrm{ff}})$, and per-step compute, $\Theta(L^2 d + L d^2 + L\,d\,d_{\mathrm{ff}})$ (identical to one Transformer layer), are both independent of $T$; unrolling for $T$ steps costs $\Theta(T(L^2 d + L d^2 + L\,d\,d_{\mathrm{ff}}))$ total FLOPs, linear in $T$, with $L$ unchanged.

Horizontal CoT, in contrast, must generate these $T$ thinking tokens one at a time, each requiring a full pass through all $N$ layers. On memory, horizontal CoT's KV-cache holds $L+T$ tokens across all $N$ layers, $\Theta((L+T)Nd)$, while our recurrence discards each step's keys and values once $H^{(t)}$ is updated and so holds $L$ tokens for a single block, $\Theta(Ld)$, independent of $T$. Their ratio, $(1+T/L)N$, separates the two sources of the saving: the $N$ is weight sharing and stays constant, while the $1+T/L$ is what emitting no tokens buys, which is independent of width, depth, and batch size, unbounded as reasoning lengthens, and correspondingly small when reasoning is short relative to the prompt.

The two paradigms also differ in \emph{latency}: since the $N$ layers within one token's forward pass are sequentially dependent, and each of the $T$ tokens must wait for the previous one to be produced, horizontal CoT's critical path spans $\Theta(TN)$ sequential layer evaluations. Our recurrence applies a single block $T$ times, so its critical path is $\Theta(T)$.

\begin{figure}[tb]
    \centering
    \includegraphics[width=.6\columnwidth]{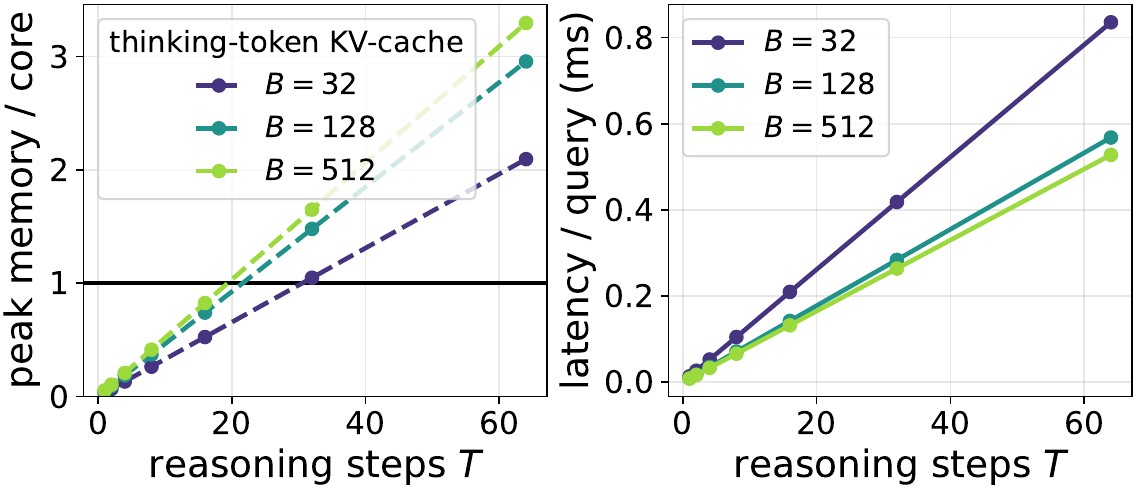}
    \caption{Measured inference cost of the reasoning core as the step count $T$ grows. Left: the key--value cache $T$ thinking tokens add to a depth-$24$, $d{=}1024$ backbone (GPT-2 medium, fp16; prompt excluded), as a multiple of the core's $T$-independent footprint (the solid line at $1$); linear in $T$, reaching $2.1$--$3.3\times$ by $T{=}64$. Right: latency per query, linear in $T$ and falling as the batch grows.}
    \label{fig:inference-cost}
\end{figure}

Figure~\ref{fig:inference-cost} confirms this scaling across serving batch sizes. The left panel compares memory. The reasoning core's peak memory does not change with $T$ at inference, so we take it as the unit (the solid line at $1$). The dashed curves show the key--value cache the backbone builds up while generating $T$ thinking tokens, measured on GPT-2 medium~\citep{radford2019language} ($N{=}24$). We count only the thinking tokens: the prompt occupies memory in both models, so it cancels out. This cache grows linearly in $T$, passing the core's entire footprint at $T{\approx}19$--$31$ and reaching $2.1$--$3.3\times$ it by $T{=}64$. The right panel reports latency per query. It is linear in $T$, as the $\Theta(T)$ critical path predicts, and falls as the batch grows: $0.84$ ms at $B{=}32$ against $0.53$ ms at $B{=}512$, both at $T{=}64$, since a wider batch amortizes each block application over more queries.

\subsection{Limitations and Future Work}
\label{sec:limitations}

\paragraph{Scale} We deliberately study reasoning cores of fewer than 1M parameters, isolating the recurrence mechanisms from large-scale pretraining confounds and keeping the depth at which each task fails measurable. The cost model of Section~\ref{sec:complexity} is independent of core width, so it carries to deployment scale. Whether identity-biased gating and LayerScale still suffice as $d$, $d_\mathrm{ff}$, and $L$ grow by orders of magnitude remains open.

\paragraph{Manually Designed Perception Interfaces} Our three interfaces (topological masking, RoPE, standard attention) were hand-selected per task; a more general system should unify these choices into one interface, e.g., a learning block that interpolates between strict priors and unconstrained attention.

\paragraph{Fixed, Externally Specified Depth} At inference time, we treat the recurrence step count $T$ as an external hyperparameter, which lets us separate depth generalization from stopping-time learning. A practical system would benefit from a lightweight halting module---in the spirit of ACT~\citep{graves2016adaptive} but applied at the sequence level rather than per token---that allocates more steps to harder instances without a preset budget. This is harder than it looks. When we train an ACT Universal Transformer, the learned halting budget is calibrated to the training distribution and does not grow with the depth that OOD instances require (Table~\ref{tab:act-ponder}).

\begin{table}[t]
\centering
\caption{Ponder steps (for graph, averaged over the source and target nodes) in the OOD regime vs.\ the depth the task requires. Low OOD accuracy coincides with a large shortfall, although the ponder count is only a proxy for propagation depth.}
\label{tab:act-ponder}
\begin{tabular}{lccc}
\toprule
\textbf{Task} & \textbf{Learned ponder} & \textbf{Steps required} & \textbf{OOD acc.} \\
\midrule
Graph (hops 8--12)   & ${\approx}4.5$  & $8$--$12$        & $50$ \\
Nested logic (d.\ 10--14) & ${\approx}10.2$ & ${\approx}10$   & $91$ \\
Family (hops 6--9)   & ${\approx}6.3$  & grows with depth & $57$ \\
\bottomrule
\end{tabular}
\end{table}

\paragraph{Integration with Pretrained LLMs} Moving the reasoning core from task-specific models to a pretrained LLM raises new design considerations:

\begin{itemize}
    \item Placement: designate a single existing layer or small contiguous block as the recurrent core and loop it in place, rather than sharing weights across \emph{all} layers.
    \item Warm-starting: initialize the gate to reproduce the pretrained layer's single-pass behavior at $T=1$, introducing extra steps gradually during fine-tuning.
    \item Parameter-efficient adaptation: leave the pretrained weights frozen and carry the recurrence entirely in a LoRA~\citep{hu2022lora} side branch, or a higher-capacity variant such as CeRA~\citep{chen2026cera}, training only the adapters with the new depth embeddings, gate, and LayerScale parameters.
    \item Trigger mechanism: decide when to invoke the recurrent block instead of unrolling it everywhere, connecting to the halting-policy question above~\citep{wei2022chain, goyal2023think, hao2024training}.
\end{itemize}

\section{Conclusion}
\label{sec:conclusion}
 
We study \emph{vertical Chain-of-Thought}: adding test-time reasoning compute by iterating a shared-weight block in latent space rather than generating tokens, so that added depth leaves peak inference memory unchanged and grows latency only linearly. We instantiate it with silent thinking, LayerScale, and an identity-biased gate, unroll it for 20+ steps, and characterize it on three compositional tasks. Test-time depth recurrence extrapolates beyond the training range: it succeeds on the graph reachability task where fixed-depth Transformers barely extrapolate, and it comes within two points of fixed-depth models several times larger on the two sequence tasks. Across the three tasks, the generalization profile varies with the perception interface, and we find that per-step supervision, a common practice for deep iterative models, undermines this extrapolation. By ``thinking deeper, not longer,'' models can spend variable computational depth without consuming the context window or the key--value cache, allowing reasoning across large query batches.

\section*{Acknowledgment and GenAI Usage Disclosure}
This work was supported by the National Science and Technology Council of Taiwan under Grant 113-2221-E-008-100-MY3. The National Center for High-Performance Computing of National Applied Research Laboratories, Taiwan, provided computational resources.
This paper was prepared with substantial assistance from Claude Code, which the author used throughout the project to implement and run experiments, draft and revise manuscript portions, and locate relevant references, which were verified against their original sources. The author formulated all research questions, determined the final claims, reviewed and approved all content, and takes full responsibility for the work.

\bibliographystyle{apalike}  
\bibliography{ref}  

\end{document}